# Machine Learning and Consumer Data

Hannah H. Chang, Singapore Management University
Anirban Mukherjee, Cornell University

## INTRODUCTION

Advances in digital technology have led to the digitization of everyday activities of billions of people around the world, generating vast amounts of data on human behavior. From what people buy, to what information they search for, to how they navigate the social, digital, and physical world, human behavior can now be measured at a scale and level of precision that human history has not witnessed before. These developments have created unprecedented opportunities for those interested in understanding observable human behavior–social scientists, businesses, and policymakers—to (re)examine theoretical and substantive questions regarding people's behavior. Moreover, technology has led to the emergence of new forms of consumer marketplace—crowdfunding (whereby entrepreneurs obtaining funds from an anonymous online crowd; Mukherjee, Chang, & Chattopadhyay 2019) and crowdsourcing (whereby organizations gather new ideas and business solutions from an anonymous online crowd; Mukherjee, Xiao, Wang, & Contractor, 2018)—which not only details people's behavior in exchange of products and services but also led to new behavior.

Making sense of the vast amount of fine-grained data about consumer behavior, however, poses nontrivial challenges for marketing researchers and practitioners. In the past, behavioral data about consumers originated from sources such as point-of-purchase scanner data, customer attitude or satisfaction surveys, consumer purchase panels, and laboratory-based experiments. These traditional sources of consumer data are much smaller in scale, much more structured (e.g., in numbers-based data formats which can be directly analyzed), and measured to purpose, than new consumer data sources. Consequently, many of the methods used to analyze traditional customer data—such as conventional econometric and statistical methods—are not designed to deal with the breadth, precision, and scale of the new consumer data sources publicly available, which tend to be unstructured—written texts, images, audios, and videos—and require parsing and processing before data can be analyzed.

Fortunately, a parallel trend to the emergence of "big data" on consumer behavior has been the emergence of computational methods and analysis techniques needed to deal with these new sources of behavioral data—which tend to be more unstructured, of much larger scale, and noisier. Specifically, data on consumers come in four basic forms: (1) structured data, (e.g., number of likes on Facebook), (2) textual data (e.g., tweets on Twitter), (3) audial data (e.g., Spotify radio advertisements), and (4) visual data (e.g., photos on TripAdvisor). Consumer data can involve only one form, such as textual messages (e.g., tweets on Twitter) and visual images (e.g., Instagram photos). Consumer data can also involve more than one form. Video data, which are increasingly prevalent, combine a series of visual images (typically, 24 visual frames per second) and an audio track. Many publicly available sources of consumer relevant data detailing people's consumption behavior involve multiple data elements. For example, consumer data from YouTube combines all four of these basic forms—audial and visual data in the video, textual data in the comments, and structured data in the number of views and likes. For each of these four data elements, new machine learning and big data methods enable us to simply and easily parse the data to uncover consumer insights if analysts are equipped with the right toolkit.

As both the availability of large-scale behavioral data and computational analysis methods are recent and emerging developments, many behavioral scientists and practitioners may be unaware or unfamiliar with (1) new sources of secondary data and types of data that are available to extracts insights about consumer behavior, and (2) new analysis techniques to study consumer behavior at scale. Therefore, motivated by these recent developments and opportunities, the main objective of this article is to discuss computational methods (specifically, machine learning methods) for researchers and practitioners interested in addressing customer-relevant questions using new secondary data sources that are publicly available. This article offers a primer on the application of computational social science for understanding consumer data for both researchers and practitioners.

The rest of this article is organized as follows. First, types of unstructured data pertaining to consumer behavior—including the information that consumers are exposed to and their digital footprints in the modern marketplace—will be decomposed to their underlying data elements. Next, machine learning and computational techniques to parse and process unstructured customer data are described. Finally, potential directions for future research using consumer data are discussed.

**TYPES OF CUSTOMER-RELEVANT DATA**

Consumers today have unprecedented access to numerous types of information and media. New forms of information environments, often fueled by technology, have also emerged since the 2010s. For example, reward-based crowdfunding platforms like Kickstarter and Indiegogo post videos and text descriptions about new product innovations (Dhanani & Mukherjee, 2017; Allon & Babich, 2020). Debt-based crowdfunding platforms such as Lending Club and Funding Societies post text descriptions about business loans (Lee, Chang, & Mukherjee, 2020). As self-contained marketplaces, crowdfunding data include comprehensive descriptions of the factors that influence consumers and investors as well as detailed accounts of purchase and exchange behavior (Mukherjee et al., 2019). App-based ecosystems enable marketers to promote "green" initiatives as a part of their corporate social responsibility strategy (Merrill, Chang, Liang, Lan and Wong 2019). These platforms offer a wealth of detailed data to study emerging consumer behavior in new marketplaces. New voice assistants such as Amazon's Alexa and Apple's Siri allows consumers to search for product information through voice based commands, simplifying consumers' purchase journey, which led to changing customer behavior. This development also led to emerging forms of voice-assisted retail shopping that is often dubbed "voice commerce" (i.e., v-commerce), all of which offer a wealth of data to study emerging customer behavior in new marketplaces. Moreover, new digital channels help facilitate communication among businesses, consumers, investors, and other stakeholders.

Much of customer-relevant data today are publicly available, giving researchers' and analysts unprecedented access to study modern consumer behavior and generate behavioral insights. For example, Yelp released an academic dataset which contains description of patrons' ratings, reviews, and images of restaurants and businesses on its platform, while Twitter offered an API that facilitates data collection of tweets posted on Twitter. Spotify also released its web API to allow easier data access for researchers to collect people's music listening behavior on the platform. Many other customer-facing firms similarly provided opportunities for researchers and analysts to access their data.

Consumers information environment—including information and media that consumers are exposed to in learning about brands and products as well as data generated by consumers' behavior—can be decomposed to four basic data elements: structured (numbers), textual, audio, and visual data. Table 1 outlines examples of customer-relevant data along the four basic data elements. They are discussed next.

**Table 1. Overview of customer-relevant data in the modern marketplace**

| Data elements | Sample sources of such data from information shown to consumers in the marketplace | Sample sources of such data captured from consumer behavior | Common computational techniques for data processing |
|---|---|---|---|
| **Structured data** | ● Price<br>● Consumer reports product ratings<br>● Product quantity<br>● Classification systems used to describe products (e.g., product categories) | ● Customer review ratings<br>● Attitude surveys<br>● Satisfaction surveys | ● Standard statistical models<br>● Standard econometrics models |
| **Textual data** | ● Product descriptions<br>● Consumer reports product reviews<br>● Packaging labels | ● Product reviews<br>● Email inquiries to customer services<br>● Social media posts (e.g., tweets on Twitter) | ● Linguistic inquiry and word count (LIWC)<br>● Natural language processing<br>● Sentiment analysis<br>● Topic modeling |
| **Audial data** | ● Audiobooks<br>● Earnings conference call<br>● Music<br>● Social audio app (e.g., Clubhouse)<br>● Sonic branding<br>● Radio<br>● Podcasts<br>● Voice assistants (e.g., Apple's Siri, Amazon's Alexa) | ● Customer service calls<br>● Conversation log with voice assistants<br>● Voice search | ● Automatic speech recognition<br>● Waveform analysis |
| **Visual data** | ● Brand logos<br>● Digital ads<br>● Print ads<br>● Product images<br>● Visual frames sampled from videos | ● Social media photos (e.g., Instagram photos)<br>● Influencer videos | ● Computer vision<br>● Image analysis and processing |

**Structured Data**

Structured data are represented by numbers across the measurement scales (nominal, ordinal, internal, and ratio scales) and are typically denoted as quantitative data. Structured data is typical and common in settings where it is feasible to directly measure the intensity and extent of communications and interactions. For example, on Tripadvisor, it is possible to measure the number of reviews that were posted for a hotel property and the rating (e.g., a 5-point scale in which five points represent highest rating) that was given by users to the hotel property, in order to understand customer engagement with the brand and property and customer satisfaction with the property. In another example, on a common crowdsourcing website (Threadless.com), users upload new designs, rate each other's designs, and purchase selected designs. On this website, it is possible to measure the ratings given by users to different designs and to relate these to sales (Mukherjee et al., 2018). Furthermore, a key source of structured data (from both online and offline sources) is data on consumers' transactions, including descriptors of what the consumer purchased, how and when the purchase was consumed, and the consumer's satisfaction with the purchase. These data frequently are the dependent variable in financial models of the firms' decision calculus such as deciding on pricing and promotions, which are used to accurately and holistically measure firms' performance (Lee et al., 2020).

**Textual Data**

Communication is an integral part of the marketplace. It fundamentally shapes the relationships, and the transmission of information, between businesses, consumers, investors, public agencies, and the society at large. Among the various types of unstructured data, extant research on communication—spanning marketing, psychology, computer science, economics, and information systems—has primarily focused on written (text-based) communication such as product reviews (e.g., Tirunillai & Tellis, 2014; Van Laer , Edson Escalas, Ludwig, & van den Hende, 2019), product descriptions (e.g., Chang & Pham, 2018; Mukherjee & Chang, 2022), movie storylines (e.g., Toubia, Iyengar, Bunnell, & Lemaire, 2019; Mukherjee et al. 2018), consumer application forms (e.g., Netzer, Lemaire, & Herzenstein, 2019), advertising slogans, social media posts (e.g., Hamilton, Schlosser, & Chen, 2017; Chang & Hung, 2018), marketing communication from businesses to consumers (e.g., Culotta & Cutler, 2016; Villarroel Ordenes et al., 2019), and newspapers or reference articles (e.g., Mikolov, Sutskever, Chen, Corrado, & Dean, 2013; Chen, Fisch, Weston, & Bordes, 2017). While these written communications are readily available on the Internet for data collection, digitization of information has further increased the availability of textual data.

**Audial Data**

Audial data represent another fruitful opportunity for understanding consumer behavior, as this type of data is becoming more common and readily available in the marketplace. On one end of the marketplace, audial data originates from businesses' audio communication with other stakeholders include audiobooks for consumers, earnings conference call to investors, music (in ads and as products), social audio app in which consumers can listen in to conversation between other experts and consumers on a specific topic (e.g., Clubhouse), radios, podcasts, voice assistants (such as Apple's Siri and Amazon's Alexa), and sonic branding (sounds that are used as brands' audio signature, such as the distinctive synthesized glissandos of THX's sound trademark "Deep Note" that is often played in the cinema). In recent years, the popularity of audio communication has grown, positively contributing to the availability of audial data. Brands like GE, Microsoft, Johnson & Johnson, and Sephora communicate their brand messages directly with customers through each brand's online podcasting ("The Message," ".future," "Innovation," and "#LIPSTORIES," respectively). Meanwhile, consumers also seem to increasingly embrace the consumption of audio information. For example, in 2008, 9% of Americans aged 12 and older reported listening to a podcast while 21% reported listening to online radio (Pew Research, 2021). In 2018, a decade later, 26% of Americans aged 12 and older reported listening to a podcast while 64% reported listening to online radio (Pew Research, 2021). Radio reaches more Americans each week (92%) than other channels—such as TV (87%), smartphone (81%), computers (54%), TV-connected devices (52%), and tablets (46%)—across all age groups (Nielsen, 2019). On the other end of the marketplace, consumers are also generating audial data such as customer service calls, voice search, and conversation log with voice assistants such as Siri and Alexa. Thus, audial data represents a promising but currently underutilized data source for extracting consumer insights (cf. Chang et al., 2021; Chang et al., 2022; Dahl, 2010; Meyers-Levy et al., 2010).

**Visual (Image) Data**

Mobile and other digital technology, as well as social media platforms, have contributed to a tremendous surge in the presence and use of digital images and digital videos. Typical examples of visual data generated by consumers include photos and videos posted by consumers on social media as well as influencer videos. Video sharing sites such as YouTube and TikTok generate exabytes of visual data each hour, most of which is archived but not systematically utilized for any form of meaningful analysis to extract behavioral insights. Consumers, particularly younger consumers, like to create, watch, and share content on these video and image sharing platforms (e.g., Instagram, TikTok) both to enrich their own lives and to create their social image. Consequently, it is now typical and common for brands to engage these consumers using videos on these websites by contracting influencers to create advertisements and product placements, and in creating and promoting content that is viewed as on point for the brand. In addition, image data produced by businesses for consumers include brand logos, digital and print ads, product visuals, and visual frames sampled from videos.

## MACHINE LEARNING METHODS FOR CUSTOMER DATA

The various types of information that customers are exposed to and in turn post digitally are typically unstructured elements such as videos, text descriptions, audio clips, and images. These various data elements are ubiquitous in the modern information environment. However, because such data are unstructured (e.g., audio, video), they are challenging to analyze. One estimate indicates that about 80% to 95% of the data that businesses have access to are unstructured (Gandomi & Haider, 2015).

Recent advancements in computational methods and analysis techniques help facilitate data processing and analysis to study customer-relevant questions. In particular, several classes of methods—natural language processing, computational linguistics, automatic speech recognition, waveform analysis, computer vision, and image analysis—are useful to parse and process unstructured elements (videos, texts, audios, visuals) that customers are exposed to and also post online (Athey and Imbens, 2019; Chang et al., 2022). Consequently, behavioral researchers and practitioners have just begun to leverage unstructured data that customers are exposed to in the marketplace (e.g., product videos posted on brands' social media accounts).

These machine learning, natural language processing, and data processing tools allow researchers and practitioners to examine consequential customer behavior at scale, leveraging new secondary data sources (as described in earlier section) and types of data (e.g., product videos in a leading online crowdfunding platform). They also allow large-scale structured data and computationally demanding unstructured data be analyzed in a fairly straightforward and cost-effective manner. These various techniques are described next.

## Big (Structured) Data

While unstructured data (such as written texts, music, and visual images) poses the challenge that they cannot be directly statistically modelled and understood mathematically, and therefore are not a good fit for conventional econometric, statistical, and psychometric methods (e.g., Lancaster, 1966; Mukherjee & Kadiyali, 2011, 2018), modern large-scale structured data is primarily difficult to deal with due to the overwhelming volume, variety, and velocity of such data. For example, consider the modern social media eco-system. In this eco-system, it is typical for consumers to generate tera-bytes of data describing a wealth of activities from the likes on comments, to shared posts, to geographical traces of where the consumer logged into the website or used the app. Each of these individual use-cases is simple to analyze for both descriptive purposes (e.g., to compute the count of likes) and for statistical modeling (e.g., to regress the count of likes on time of day to work out when a post is most likely to be liked). To facilitate the training of a statistical learning model using large-scale structured data, researchers can leverage established machine learning methods—supervised (e.g., random forest, LASSO regressions), unsupervised (e.g., clustering, autoencoders), and semi-supervised learning (e.g., transductive support vector machine)—to help make predictions for data-driven decisions (see Athey & Imbens, 2019 for a discussion on recent machine learning methods for empirical research and econometrics). With sufficient data for model training, these methods are straightforward to implement as there are many readily available packages in Python and R to simplify the implementation of these popular machine learning algorithms.

Often, the key analytical challenge with large-scale structured data is to determine how to organize the data analysis such that the data can be collected, stored, and analyzed in a principled workflow (see Lazer et al., 2009). Fortunately, the rise of cloud services provides both managers and researchers with the requisite toolkit. Specifically, services like Google Cloud Storage offer space in the cloud that allows for the storage and retrieval of large-scale data while other cloud services such as Google BigQuery enable such activities to be paired with software such as Tableau and AutoML to enable analysis without requiring the researcher or practitioner to know how to write code. Furthermore, modern analysis models such as neural networks, as implemented in software frameworks such as PyTorch and TensorFlow, enable the analyst to "let the data speak" – they enable relaxing functional form restrictions and the discovery of higher-level structure in high dimensional data.

**Textual Data: Text Mining And Natural Language Processing**

Unlike structured data, textual data is unstructured and needs to be converted into structured, numbers-based formats for further analysis. Typically, text mining is used initially to collect and convert unstructured textual data (e.g., text descriptions of a product innovation on a crowdfunding webpage; Younkin & Kuppuswamy, 2018) into structured formats for subsequent processing and analysis. Numerous models and approaches have been developed to analyze textual data, typically with the following aims: (1) to extract meaning of a word (or n-gram) based on count, co-location, or co-occurrence (e.g., "entity extraction" via using human-validated dictionaries and classification tools via pre-determined rules or machine learning; e.g., Pennebaker et al., 2007); (2) to extract higher-level, general themes shared across the texts (e.g., latent semantic analysis, topic modelling, and poisson factorization; e.g., see Toubia et al., 2019; Toubia, 2021); (3) to identify higher-order relations among extracted words or entities (e.g., "relation extraction" through deep learning and embedding; e.g., Toubia & Netzer, 2017).

Today, the term "text analysis" is often used synonymously with natural language processing (NLP), a class of techniques aimed to understand human languages. The original natural language processing toolkit for text analysis was relatively simple and rudimentary and primarily based on word counts, which simply counts words of different types to construct measures such as the extent of positivity and negativity in the text (e.g., sentiment analysis). These models often rely on dictionaries that are generated and validated by human participants, whose inputs are then used to infer the psychological constructs. Research in computational linguistics has led to development of statistical models that link the use of words and phrases in a communication to higher-order psychological constructs such as agreeableness in the message sender and recipient (Fast & Funder 2008; Pennebaker et al., 2003). A classic linguistic-analysis model builds on a seminal method by Pennebaker et al. (2001) that is commercially available as the Linguistic Inquiry and Word Count (LIWC; Pennebaker et al., 2015). LIWC has been widely used across disciplines, such as psychology, communications, and marketing, to study the relationship between linguistic characteristics of texts and individuals' behavior, such as in consumption (e.g., Cavanaugh et al., 2015) or lending (e.g., Netzer et al., 2019). In addition to LIWC, many other dictionary-based tools are readily available for application.

These methods have now been supplanted by the use of machine learning and deep learning (neural networks based) models that use more advanced techniques such as Transformers (Vaswani et al., 2017) to provide a much more fine-grained analysis of the text. Deep-learning (neural networks) based language models take into account the context—that is, the surrounding words to the focal word (e.g., through co-occurrence) and the remaining texts in the documents—to derive meaning of the focal word (Mikolov et al., 2013; Mukherjee & Chang, 2022). Moreover, in the last few years research has shown that pairing context-sensitive word embeddings with neural network (deep learning) models for sequential data (e.g. recurrent neural networks) leads to significant improvements in downstream performance on tasks as diverse as language inference and paraphrasing (see Peters et al., 2018; Devlin et al., 2018). Researchers can select among a wide range of NLP methods to train their context-specific language models using the textual data they have on hand, if they have sufficient data. Existing tools are available as suites of libraries and programs on Python and R, such as the popular Natural Language Toolkit (NLTK) for Python . Alternatively, many cloud service providers (e.g., IBM Watson, Google cloud, Amazon cloud, among others) and commercially available services (e.g., Hedonometer, Diction 5.0) offer pre-trained language models that are developed using existing text corpora (e.g., Wikipedia, Google news) and already validated by human participants. For example, IBM Watson offers two versions of its linguistic-analysis model, a version trained using general text documents and another trained using extant customer service dialogs. The linguistic-analysis model takes as input several lexical features commonly used in computational linguistics, including lexical features from LIWC (among other dictionaries developed and validated by human participants). IBM's linguistic-analysis model comprises of Support-Vector Machine models (a machine learning model) that were trained independently to classify each tone using a One-Versus-Rest strategy on large corpora of documents. During prediction, the linguistic-analysis model identifies the tones that are predicted with at least 0.5 probability as the final output. As a result, the model measures tones corresponding to language ("Analytical," "Confident," and "Tentative"), social propensity ("Agreeableness," "Conscientiousness," "Emotional Range," "Extraversion," and "Openness"), and emotion ("Anger," "Disgust," "Fear," "Joy," and "Sadness"; see Kaminski 2017). The wide range of readily available toolkits make analyzing large amounts of textual data much more accessible to researchers and analysts.

**Audial Data: Automatic Speech Recognition and Waveform Analysis**

Audial data in the marketplace can be decomposed to spoken narration (verbal content), voice conveying the narration (nonverbal content), music, and background sound. This type of unstructured data is typically captured in acoustic waveform in its raw form. To parse and process audial data, researchers and practitioners need to first identify the research focus, specifying the audial features to characterize (for feature extraction). For example, podcast advertising typically includes a voiceover narration—narrators' voices conveying spoken content—music and other sounds (Chang et al. 2022). These different audial components require different data processing techniques prior to data analyses. Researchers dealing with spoken narration can apply automatic speech recognition (ASR), a process by which a computer system maps acoustic speech signals to texts. ASR generates text transcription of the spoken verbal content and deduces speaker identity as model outputs.

Intuitively, speech recognition models involve pattern recognition through speech detection, segmentation, and clustering of spoken utterances. ASRs operate through multiple phases. First, the model detects which audial patterns in the waveform is speech. Next, the model detects when speakers in the audial utterances change by examining variations in the acoustic spectrum (through vocal features such as pitch and timbre). Finally, the model identifies what was said by each speaker in the entire acoustic waveform, through audial features characterizing each speaker. Thus, through a procedure known as speaker diarization via sequence transduction, ASRs operate on the acoustic waveform from which they extract the data features attributed to the identity of different speakers and to identify the words spoken by each speaker in the audio track.

The current state-of-the-art speech recognition model leverages deep learning (in particular, recurrent neural network (RNN)) methods to accomplish these tasks, automating transcription of audial speech data to text-based data (Graves, Mohamed, & Hinton, 2013).

Whereas non-machine-learning-based ASRs use gaussian mixture models, hidden markov models, or their combination to map speech fragments with audial bits, modern ASRs use a much larger latent state-spaces to capture long-term dependencies in sequential data such as speech, given that the focal spoken word depends on what was said before and what was said after. Because deep neural networks such as RNNs are able to infer the word that is spoken from both the sequence of spoken syllables and the sequence of spoken words, deep-learning-based speech recognition models can achieve greater computational efficiency and accuracy than conventional speech recognition models (Chelba et al., 2013). ASRs are relatively mature in their development among the recent techniques for unstructured data, with many commercially available toolkits. Some of the most popular ones include speech-to-text APIs and commercially available services from Speechmatics, Otter.ai, and Google's cloud services. Given that a main output from ASRs is the text transcription of spoken content in the acoustic waveform, researchers can then use text analysis and natural language processing methods (discussed in the earlier section on textual data) to treat the resulting textual data for analysis of spoken content.

Besides spoken content, researchers and analysts may be interested in deriving acoustic characteristics of speech, music, and background sounds in the raw audio waveform. Typically, audial characteristics are described by (1) physical features (e.g., spectral features, volume), (2) perceptual features (e.g., pitch), and (3) signal features (e.g., zero-crossing rate; Fant, 1960). Numerous toolkits are readily are available to help measure audial characteristics, based on researchers' and analysts' focus of audial elements. For example, many tools are accessible derive acoustic features of songs and music, such as sonic API, Discogs API, Spotify API, and Last.fm API. To measure more general acoustic features of audio waveforms, toolkits such as the Librosa library in Python and NVIDIA's Data Loading Library (DALI) API (for Python and PyTorch) can help derive physical features characterizing the audio spectrogram (i.e., a representation of an audio signal of the frequency spectrum over time) such as short-time fast fourier transform (the process of dividing the audio signal in short term sequences of certain sizes and computing these) and spectral centroid (location of the center of mass of the spectrum). These toolkits can also help derive signal features such as zero-crossing rate (a measure of the rate at which the signal is crossing the zero line between negative and positive signal; Chang et al. 2021).

**Visual Data: Computer Vision And Image Analysis**

Visual images are unstructured and require processing to represent them in numbers for statistical analyses. An image is essentially a 2D or 3D function of spatial coordinates, with its amplitude at a particular value of spatial coordinates determining the intensity of an image at that point. An image is represented by an array of pixels arranged in vectors, where pixels are elements that contain information about intensity and color. Basic image analysis methods can help measure pixel-by-pixel variation to account for data features of visual characteristics, such as variation in visual imagery (without knowing the visual content and objects in the image) and resolution (number of pixels in the image, which is linked to image detail). These methods can also help transform an image by varying these data elements of a visual image.

Recent developments in new algorithms, digital images, and computing power have jointly contributed to new machine learning methods which allow computer systems to detect, identify, and understand visual images. A popular class of machine learning methods typically used to process digital images is computer vision. Broadly speaking, computer vision concepts constitute three levels of techniques, with varying focus on data outputs (and increasing sophistication) for each level. Low level computer vision techniques derive basic data features of the image content, extracting fundamental image primitives such as edge detection (by detecting discontinuities in brightness), corner detection, shapes, morphology, and filtering. It is typically used to preprocess images. Middle level computer vision aims to infer visual geometry and motion, which require computer systems to compare more than one images. Finally, high level computer vision techniques attempt to understand the "semantics"—the content of visual images—such as object recognition and scene understanding. In other words, higher level computer vision techniques attempt to mimic human visual systems and higher-order cognitive processes and are computationally more intensive.

In higher level computer vision, deep-learning-based visual recognition models typically use convolutional neural networks (CNNs) that are trained on large corpora of labelled images (such as ImageNet) to derive the data features (e.g., colors, lines, and curves) corresponding to images of common objects (e.g., tables, humans, cats; Krizhevsky, Sutskever, & Hinton, 2012), which allow computers to make probabilistic inferences on the visual contents of the images.

There are many readily available open-source libraries for Python and R, as well as commercially available services, with built-in algorithms and pre-trained models for common computer vision tasks. Among open-source libraries (for Python or R), OpenCV and Scipy are useful for a wide range of lower and higher level computer vision tasks such as object recognition and face detection, whereas Python image library (PIL), FFmpeg, and Numpy are useful for lower level image processing tasks such as filtering, measuring of pixels, and manipulating images. Other open source APIs, such as NVIDIA's DALI API, can also help carry out common image processing tasks. Higher level computer vision tasks such as object identification and face recognition can be completed using commercially available services, Clarifai, Google cloud services, and Lionbridge AI. For example, researchers and analysts can leverage a visual-recognition model developed by Clarifai (a state-of-the-art CNN architecture trained over billions of training samples) to identify the visual elements (common objects) present in each image frame. The model operates automatically on the images and outputs a vector of words describing each frame (e.g., tables, people, cats, etc.). These automated image analysis techniques can help parse and process image data for subsequent data analysis using standard statistical and econometrics models (see Chang et al., 2022).

## FUTURE RESEARCH DIRECTIONS

Information display and media can combine more than one data element. For example, a print ad can include an image of the product and a text-based message about the product. Video is another form of audio-visual data. For decades, companies have relied on television commercials to communicate their brands and products to consumers. Advances in internet and digital technology have added to the popularity of newer forms of marketing communications carrying audio-visual content, such as product videos (e.g., new product introduction or product review videos) and branded vlogs. Consumers also add to the availability of video data by posting short videos on social media.

A video is composed of an audio track and a sequence of images (frames). Broadly speaking, video analysis is a combination of analyzing audio track and visual images. To analyze videos, researchers and analysts need to first decouple the audio track from the visual frames. This can be easily done using any of the image analysis libraries, APIs, and commercial services (for lower level computer vision techniques) discussed earlier. Depending on the length of the video, it may be necessary to sample the visual frames as a video typically has 24 frames per seconds and analyzing each visual frame is computationally more costly than standard data analysis of numbers. A series of deep-learning models (in automatic speech recognition, computational linguistics, and computer vision) can then be applied to the audio tracks and visual frames, which allows researchers to parse the unstructured multimedia data and measure the focal and control variables for empirical analyses (e.g., Chang et al., 2022). As video analysis is one of the exciting areas receiving much research attention in computer science and artificial intelligence, it is likely that the current empirical challenge with detailed analysis of large-scale video datasets would be solved in the foreseeable future.

## CONCLUSION

This article presents an overview of new sources of customer-relevant data that are publicly available—such as purchase behavior in popular crowdfunding platforms and social media posts (texts, images, videos). These data sources present fruitful opportunities for researchers and practitioners to better understand modern consumer behavior in the evolving information environments. Not only would resulting insights be theoretically interesting, they can also offer practical benefits for extracting insights about customer behavior and for designing the information environment that the modern consumers face. Finally, the availability of new types of secondary data and the use of machine learning methods create new fruitful avenues to study emerging consumer behavior. These tools allow us to examine consequential consumer behavior in the marketplace at scale, leveraging new secondary data sources and types of data. Machine learning methods facilitate data processing and analyses of data sources that were challenging to analyze just a few years ago, leading to new possibilities for researchers and analysts to extract novel behavioral insights about individuals and groups at scale. These developments create exciting opportunities for behavioral researchers, social scientists, practitioners, and businesses going forward.

## ACKNOWLEDGEMENT

This research was supported by the Ministry of Education (MOE), Singapore, under its Academic Research Fund (AcRF) Tier 2 Grant, No. MOE2018-T2-1-181. Any opinions, findings and conclusions or recommendations expressed in this material are those of the authors and do not reflect the views of the Ministry of Education, Singapore.

## ADDITIONAL READINGS